\documentclass[manuscript,screen]{acmart}
\usepackage{graphicx} 
\usepackage{amsmath}
\usepackage{algpseudocode}
\usepackage{algorithm}

\DeclareMathOperator*{\argmin}{arg\,min}

\date{November 2023}

\begin{document}

\title{Longitudinal Counterfactual Explanations: Constraints and Opportunities}

\author{Alexander Asemota}
\email{alexander.asemota@berkeley.edu}
\affiliation{
    \institution{University of California, Berkeley}
    \city{Berkeley}
    \country{United States}
}

\author{Giles Hooker}
\email{ghooker@wharton.penn.edu}
\affiliation{
    \institution{University of Pennsylvania}
    \city{Philadelphia}
    \country{United States}
}

\begin{abstract}
    Counterfactual explanations are a common approach to providing recourse to data subjects. However, current methodology can produce counterfactuals that cannot be achieved by the subject, making the use of counterfactuals for recourse difficult to justify in practice. Though there is agreement that plausibility is an important quality when using counterfactuals for algorithmic recourse, ground truth plausibility continues to be difficult to quantify. In this paper, we propose using longitudinal data to assess and improve plausibility in counterfactuals. In particular, we develop a metric that compares longitudinal differences to counterfactual differences, allowing us to evaluate how similar a counterfactual is to prior observed changes. Furthermore, we use this metric to generate plausible counterfactuals. Finally, we discuss some of the inherent difficulties of using counterfactuals for recourse. 
\end{abstract}

\maketitle

\section{Introduction}

Over the past two decades, machine learning and artificial intelligence have become intertwined with broad swaths of society, from education to criminal justice to consumer finance. Throughout this transition away from human decision-makers and towards algorithmic decision-makers, researchers, practitioners, and advocates have emphasized the need for explainability and transparency. Approaches to explainability have varied widely, from creating novel 'glass-box' model architectures to developing post-hoc local explainability techniques \cite{ebm} \cite{lime}. Of particular interest in the past five years are \textit{counterfactual explanations}, which explain an individual prediction by finding an, in some sense, small change to achieve a desired prediction\cite{counterfactuals}. In contrast to most explainability techniques, counterfactual explanations seek to explain algorithmic decisions to data subjects. 

Although there has been substantial work in the domain of machine learning explainability, significant gaps exist regarding the utility of explanations to data subjects. Unlike concepts such as accuracy or sparsity, subject utility has neither a simple nor agreed upon mathematical definition. Consequently, counterfactual explanation methods optimize subject utility using disparate approaches \cite{face} \cite{dice}. Terms such as plausibility, validity, and actionability are used to describe different aspects of the utility of counterfactuals. Plausibility, the main focus of this paper, requires that a counterfactual is a possible state of being \cite{lit_review}. Nonetheless, generating plausible counterfactuals is not a simple task. Substantial effort has been devoted to developing methods for plausible counterfactuals, but there are no agreed upon approaches or even metrics for plausibility. Additional effort has gone into using counterfactuals to provide recourse to data subjects \cite{karimi20_recourse}. Recourse is a stricter goal, requiring that a counterfactual be useful to a data subject in pursuing a desired decision. Therefore, plausibility is necessary for counterfactuals to be used for recourse. 

This paper proposes a novel approach to evaluating plausibility using longitudinal data. We begin by briefly reviewing approaches to improving plausibility in counterfactuals, discussing in particular persistent pitfalls. We then introduce a longitudinal distance metric for counterfactual explanations. In introducing our metric, we bring forth the benefits of using longitudinal data as a proxy for plausibility and mention some limitations. Next, we perform experiments with our metric to evaluate the use of longitudinal data for plausibility. We also explore some of the consequences of requiring plausibility. Finally, we discuss the implications of our results in the broader context of providing recourse through counterfactual explanations. 

\section{Counterfactual Explanations and Recourse} \label{cfe}

We begin by formulating the problem of counterfactual explanations. We borrow the following framework from \cite{diversity}.

Suppose we have a predictive model $f : \mathcal{X} \rightarrow \mathcal{Y}$, where $\mathcal{X}$ is the feature space and $\mathcal{Y}$ is the outcome, usually with $\mathcal{Y} = \{0, 1\}$. Then, for some given example $x$, a counterfactual explanation can be derived from the following:
\begin{equation}\label{pen_cf}
    e^* = \argmin_{e \in \mathcal{E}} pen_x (e)
\end{equation}
where $pen_x(e)$ is a penalty term that depends on $x$ and $\mathcal{E}$ is the set of 'acceptable' counterfactuals. Both $pen_x(\cdot)$ and $\mathcal{E}$ vary significantly between methods, but they often correspond to a measure of distance from $x$ to $e$ and data points with the desired label, respectively. This setup is deliberately broad, but largely captures how different methods approach counterfactuals. For example, Wachter et al., who coined the term 'counterfactual explanation', define $pen_x(e) = \Vert x - e \Vert_1$ and $\mathcal{E} = \{e \in \mathcal{X} | f(e) \approx y'\}$, where $y'$ is a desired prediction \cite{counterfactuals}. Though Wachter et al. solve the problem using Lagrangian multipliers, \ref{pen_cf} captures their approach. 

With this framework in mind, we can differentiate counterfactual explanation methods along a few axes:
\begin{enumerate}
    \item Definition of $pen_x(\cdot)$
    \item Determination of $\mathcal{E}$
    \item Optimization of $\argmin_{e \in \mathcal{E}} pen_x (e)$
\end{enumerate}
These three characteristics largely discriminate between methodologies. 1 and 2 in particular are relevant for our discussion of plausibility in counterfactual explanations. To produce plausible counterfactual explanations, we must constrain the search space of counterfactuals to the set of plausible counterfactuals. We can achieve this by penalizing implausible counterfactuals with $pen_x(\cdot)$ or by only considering plausible counterfactuals in $\mathcal{E}$. Both approaches have benefits and limitations. When introducing a plausibility term in $pen_x(\cdot)$, our optimization may wander into spaces that have no plausible counterfactuals, but we avoid having to decide exact constraints. In contrast, if we attempt to only include plausible counterfactuals in $\mathcal{E}$, we have to make a decision about which features should be constrained and how, though we can be certain that generated counterfactuals will be plausible. 

\subsection{Offering Recourse Through Counterfactuals}

A common motivation for counterfactual explanations is to provide data subjects with a path to recourse. Counterfactuals are unique in their ability to not only explain algorithmic decisions to a lay audience, but also explain how someone could receive a desired decision. That is, a counterfactual informs a subject not only \textit{why} they received a decision, but \textit{what} to change and \textit{how much} to change. Therefore, counterfactuals have the potential to greatly increase transparency and accountability in algorithmic decision-making.

However, persistent gaps exist between the ideal scenario and counterfactuals in practice. Centrally, current methodology fails to consistently produce plausible or achievable explanations. Here, we use the terms plausible and achievable to refer to \textit{objective} and \textit{subjective} perspectives of the difficulty of pursuing a given counterfactual. A counterfactual is \textit{plausible} if it respects constraints on reality, for example, not changing ethnicity or decreasing age. On the other hand, a counterfactual is \textit{achievable} if the relevant subject can achieve it. It is generally plausible for someone to increase their level of education, but it may not be achievable for a given individual. These definitions themselves elucidate the difficulty of offering recourse through counterfactual explanations; how do we know if a data subject can act on a particular recommendation? Existing counterfactual explanation methods use proxies for plausibility and achievability in an attempt to avoid implausible recommendations. 

A common technique is simply to ask the user (i.e. the person using counterfactuals to explain an algorithm) for constraints on features \cite{dice} \cite{geco} \cite{face}. Counterfactual explanation methods often ask which features should be changed or the range of allowed changes before generating counterfactuals. Then during generation, the counterfactual optimization scheme will only search within the specified bounds. Asking for user constraints has multiple purposes, but this is most often proposed to improve plausibility. A user's subjective beliefs about the mutability of features are valuable for producing plausible counterfactuals, so we can, in theory, rely on their judgement when deciding how to constrain our search space. User constraints can be effective when the user has a strong grasp of their data and the context of data subjects. However, this approach is largely limited to the knowledge of the user. In practice, there may be too many features for a user to make a principled decision, or the user may simply have incomplete information of the prediction task. Social bias also can easily leak into this approach. 

Other methods seek to use structure in the data as a proxy for plausibility. This approach comes in two forms, the more common of which is to look for counterfactuals near observed data \cite{moc} \cite{cruds}. Broadly speaking, the density of data informs us of the likelihood that a counterfactual is plausible. If a counterfactual is near prior observed data, then we can reason that the counterfactual is a possible state of being. Therefore, we can improve plausibility by requiring counterfactuals be near observed data. This helps us ensure that a counterfactual is something we could potentially observe in reality, instead of containing feature combinations that are never observed. Enforcing density constraints is a straightforward solution, however, data density may hide important factors. For example, if ethnicity is included in the predictive model, there may be higher density along the ethnic majority, leading to worse explanations for ethnic minorities. Moreover, recourse requires that a counterfactual be achievable for a specific individual, and density constraints only assist in verifying that a counterfactual could be some observed person. 

The second approach that leverages structure in data uses causal graphs \cite{karimi20_recourse}. Causal graphs show how random variables are causally connected to each other, which could in turn be used to generate counterfactuals that respect those connections. Similar to density constraints, respecting causality can prevent generation of counterfactuals that contain implausible combinations of features. Moreover, causal constraints are stronger than density constraints. As such, causal graphs are powerful in theory, especially in dealing with correlated features and downstream effects. However, they require significant domain knowledge that often is not available or robust enough to constrain decision-making. Causal inference is a challenging problem that often requires close inspection for one or two variables, so determining causality between dozens or hundreds of features is largely intractable. 

Ultimately, proxies are needed to produce plausible or achievable counterfactuals at scale. We may not know what an individual can or cannot achieve, or we may have incomplete information of relationships between the features in our model. However, existing methodologies often use proxies that insufficiently penalize implausible explanations. Particularly, there are gaps between what existing methods seek to penalize and what they actually penalize. Though using proxies often entails some disconnect between intention and action, the current state of proxies for plausibility prevents the use of counterfactuals explanations for recourse.

\section{Longitudinal Data as a Proxy for Plausibility} \label{longitudinal}

As discussed in \ref{cfe}, counterfactual explanations often are motivated by the goal of offering recourse to data subjects, though there are persistent issues that prevent most methods from providing recourse. If we view counterfactual explanations as potential paths to recourse, then we can conceptualize them as recommendations for algorithmic subjects. Specifically, we can view counterfactuals as recommendations for changes that a data subject can make to receive a desired decision at some point in the future. Conceptualizing counterfactuals as potential states of being forward in time naturally leads to considering longitudinal likelihoods. That is, when making recommendations for the future, we should consider prior observed changes over time. This perspective leads us to the primary goal of this paper: leveraging longitudinal data to assess and improve plausibility in counterfactual explanations. 

\subsection{Methods}

\subsubsection{A Longitudinal Distance Metric}\label{metric}

We introduce a distance metric that compares prior observed changes to proposed changes in the form of counterfactual explanations. 
Let $A, B \in \mathbb{R}^{n \times d}$ be $n$ observations of $d$ features across two different points in time. Subsequently, let $D = B - A$, that is the change in the observed features over time. Now we define our distance metric
\begin{equation} \label{distance}
    L(x, e; D, s) =  \min_{|\mathcal{I}| = s} \frac{1}{s}\sum_{i \in \mathcal{I}} \|(e-x) - D_i\|
\end{equation}
where $\mathcal{I}$ is an index set for prior observed changes and $s$ is the desired size of the index set. In summary, we compare a proposed difference to the $s$ closest differences and average them. 

We can justify and augment our approach in the following ways:
\begin{itemize}
    \item Since there are likely a large variety in observed trajectories, we average the $s$ most similar. This allows room for heterogeneity across trajectories without allowing a single rare trajectory to dominate our metric.
    \item In the likely chance that our data contains heterogeneous features, we can normalize our distance metric across features. Here, we consider dividing features by a metric for the dispersion of their observed differences. Our experiments use the median absolute deviation (MAD) or average absolute deviation (AAD), but other approaches can be implemented.
    \item For categorical features, several options can be exercised. For binary features, the average absolute deviation can be appropriate. For multi-class features, we normalize by the rate at which changes are observed in the longitudinal data. 
    \item Normalization can empower discovery of implausible counterfactual explanations. If our feature has a high normalization value (e.g. 1 over the MAD), then changes to that feature are rarely seen. 
\end{itemize}

Our proposed metric is flexible in its use; we can use it both during and after generating counterfactuals. Post-generation, the longitudinal distance metric can be used to evaluate and rank the plausibility of explanations. During generation, the distance metric can be used to further constrain the search space. Notice that in comparing proposed changes to those observed in longitudinal data, we not only re-weight distances on a per-feature basis, but we also incorporate dependencies between feature changes; ruling out a requirement to, for example, both change profession and increase the length of tenure in your current job. 

\subsubsection{Scoring Plausibility in Counterfactuals}

A first approach to incorporating our longitudinal metric is to re-score a proposed collection of counterfactuals. Stochastic search algorithms used in \cite{dice} and \cite{moc} return a set of possible counterfactual explanations, usually incorporating a geometric distance metric; these can then be examined or prioritized by longitudinal distance. We generate counterfactuals using the methods in \cite{dice}, and then score them by plausibility. This two-step approach allows us to use the more regular geometric distance for optimization, providing a more efficient search of feature space. In this paper, it also allows us to examine the baseline plausibility of counterfactuals generated with a geometric distance. 

\subsubsection{Genetic Longitudinal Counterfactuals}

As mentioned in \ref{metric}, our longitudinal distance metric can be used during generation as an additional constraint on counterfactuals. While our metric can be included in a multitude of ways, we propose jointly optimizing a longitudinal objective and a proximity objective (defined by Equation \ref{proximity}). Inspired by other counterfactual methods that optimize multiple objectives, we use a genetic algorithm to generate counterfactuals constrained by longitudinal data. As a point of comparison, DiCE-genetic minimizes a proximity objective and a sparsity objective (defined by Equation \ref{sparsity}). 

\begin{equation} \label{proximity}
    L_{prox}(x, e) = \left(\sum_{i \in Continuous}\frac{1}{MAD(X_i)}|x_i - e_i|\right) + \left(\sum_{i \in Categorical}(x_i \neq e_i)\right)
\end{equation}

\begin{equation} \label{sparsity}
    L_{sparse}(x, e) = \sum_i (x_i \neq e_i)
\end{equation}

In the genetic algorithm, largely borrowed from \cite{dice} and \cite{geco}, we begin by generating a random population of the desired class. Then, we assess the fitness of the population relative to our input and rank the population by fitness. The top half of the population is then mated (i.e. features are randomly chosen between two individuals). The next generation in the algorithm is made up of the top half of the current generation and their offspring. We repeat this cycle until the best fitness does not change substantially. 

Algorithm \ref{alg:cap} provides pseudocode for the above description. This algorithm is flexible enough to allow for a variety of fitness metrics, and in this paper we use our longitudinal metric to generate counterfactuals constrained by longitudinal distances. 

\begin{algorithm} 
\caption{Genetic Counterfactuals}\label{alg:cap}
\begin{algorithmic}
    \State \textbf{Input:} (subject input $x$, desired outcome $z$, model $M$)
    \State POP $\gets$ \textbf{IntialPopulation}($x, z$)
    \State currentBest $\gets \infty$
    \Repeat 
        \State prevBest $\gets$ currentBest
        \State mostFit $\gets$ \textbf{SelectFittest}($x, z, M$) 
        \State currnetBest $\gets$ \textbf{BestFitness}(mostFit)
        \State POP $\gets$ \textbf{Mate}(mostFit)
    \Until{currentBest $\approx$ prevBest}
    \State \textbf{return} POP
\end{algorithmic}
\end{algorithm}

\subsection{Benefits and Limitations of Our Approach}

Our approach is motivated by the conceptual benefit of leveraging longitudinal data, but practical benefits are equally present. Primarily, longitudinal data provides us with something close to the 'ground truth' of plausibility for data subjects seeking recourse. While still a proxy for our ultimate goal of counterfactual plausibility and achievability, longitudinal data offers some significant benefits over existing proxies. Additionally, longitudinal data allows us to assess plausibility without input from the user. As such, we can avoid some of the pitfalls that shortsightedness on the part of the user may cause. Leveraging longitudinal data shares some similarities with leveraging data density, but there are some important distinctions. Primarily, we consider not if a counterfactual is near observed data, but if the change a counterfactual implies is similar to prior observed changes. This is crucial; to use counterfactuals for recourse, we must consider a counterfactual to be an extension of the relevant data subject, not a random new person in the population. In other words, to offer recourse, a counterfactual cannot be something that someone could be, the counterfactual must be something that a specific data subject could be. 

Although longitudinal data offers an improvement over prior proxies for plausibility, there are still limitations. Centrally, longitudinal data tells us what changes have previously occurred, but it does not inform us to what extent changes were performed intentionally. This limitation is important to keep in mind; a change should not be recommended simply because it has been observed before. Recommendations may also lead to different longitudinal distributions than observing changes without intervening. Finally, longitudinal data has the same limitations as cross-sectional data; missing, unrepresentative, and noisy longitudinal data could misdirect counterfactuals if not properly analyzed. 

\section{Explorations in Evaluating plausibility} \label{exp}

In this section, we explore using longitudinal data to assess and improve plausibility for counterfactuals. We have two primary goals with our experiments. First, we want to show that longitudinal data can offer important insight on plausibility. Second, we want to evaluate the difficulty of generating plausible counterfactual explanations. 

\subsection{Ranking Explanations from DiCE and MIMIC-III} 
\subsubsection{Setup}

First, we consider ranking explanations using our longitudinal distance metric. To explore this approach, we train a model then generate counterfactuals for a held-out test set. We subsequently evaluate counterfactuals by their longitudinal distance value. 

In this experiment, we use MIMIC-III, an electronic health records (EHR) dataset, to predict acute respiratory failure (ARF) within four hours of admission \cite{mimic} \cite{mimic_data}. Specifically, we use a version of MIMIC-III that has been preprocessed by FIDDLE, an EHR data processing pipeline \cite{fiddle} \cite{fiddle_data}. Our longitudinal data consists of measurements when the patient is admitted and repeat measurements four hours after admission. We train a random forests model to predict ARF using only the first time step of data, and use DiCE (\cite{dice}) to generate ten counterfactuals for individuals in the test set who are predicted to have ARF. Our dataset contains 1350 features to train our model, twenty of which are derivations from vital signs. 

To rank counterfactuals, we use the longitudinal differences between the first and fourth hour of data. We normalize our metric using the AAD of the longitudinal differences, and we add a small tolerance ($10^{-5}$) to prevent division by zero. Finally, we conduct this experiment in two different settings: allowing all features to be changed (\textit{ALL}) and allowing only vital signs to be changed (\textit{VITAL}). These two settings should show us how our longitudinal distance metric can help assess plausibility, both when we know what features can be changed and when we may be unsure how to constrain the counterfactual search space. With this experiment, we seek to assess the discriminatory ability of our distance metric and subsequently evaluate plausibility for counterfactual explanations. 

\subsubsection{Results}

\begin{figure}[t] 
    \includegraphics[scale=0.4]{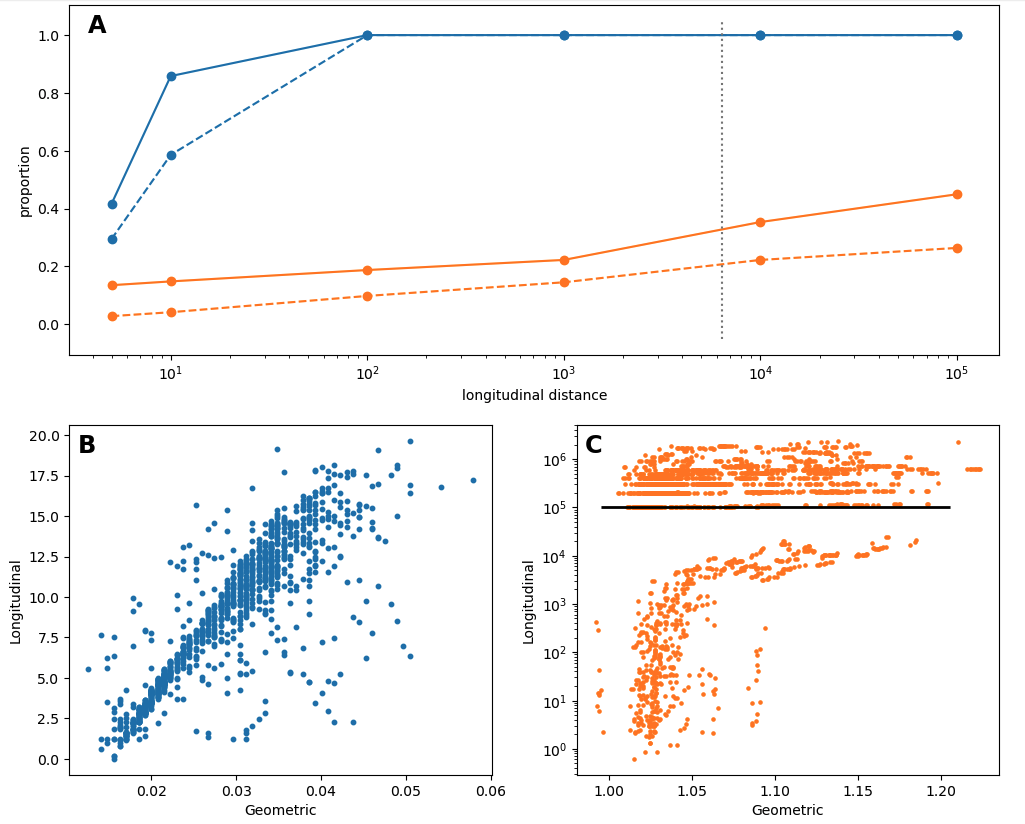}
    \centering
    \caption{\textbf{A} shows the proportion of individuals with explanations below each threshold of longitudinal distance (x-axis). The solid lines represent the proportion of individuals with at least once explanation below the threshold, and the dotted line is the average proportion of explanations below the threshold. The vertical dotted line shows the distance value we would expect a feature to have if we observed it changing for one individual. \textbf{B} and \textbf{C} compare the Geometric and Longitudinal distances between an input and a counterfactual. In all plots, blue refers to \textit{VITAL} and orange refers to \textit{ALL}.}
    \label{mimic_plot}
\end{figure}

We begin by looking at the relationship between the geometric distance and longitudinal distance. Figure \ref{mimic_plot} B and C plot the L1 distance compared to the longitudinal distance for explanations using only vital signs and all features respectively. Overall, our metric is loosely correlated with the L1 distance, but there are significant deviations based on which features are changed. In the case of explanations that can only change vital signs, there is a noticeable linear relationship. Vital signs in our dataset generally change at a similar rate, so the cost of changing one or the other is similar. Therefore, the L1 distance is closely related to the longitudinal distance. 

Looking at explanations that consider all features, the relationship is much more tenuous. Some features rarely change, leading to significant jumps in our distance metric even when only one feature is changed. Additionally, about half of the explanations change some immutable feature, such as Hospital Ward or Religion. Some explanations also change features that are mutable in theory, but not observed to have changed. Since the AAD is zero for some features, changing those features results in a large distance value. The resulting distance is above $10^5$, that is $\frac{1}{AAD+tolerance}=10^5$. Moreover, changing features without observed changes is not a rare occurrence. When we allowed any feature to be changed, 74 percent of the counterfactuals generated had a longitudinal distance value above $10^5$. This is in stark contrast \textit{VITAL}, where there are no explanations with a longitudinal distance value above 20. The disparity in longitudinal distance between \textit{VITAL} and \textit{ALL} is related to the number of features changed; in \textit{VITAL} an average of 5 features are change compared to 30 for \textit{ALL}. However, we would expect the average longitudinal distance to be much lower for \textit{ALL} if all features were as interchangeable as vital signs. Instead, counterfactuals for \textit{ALL} change features that change less often in addition to changing more features, leading to much higher longitudinal distance values.

Next, we consider plausibility at the individual level. Each individual receives ten counterfactuals, but our metric will help us see how many plausible counterfactuals an individual receives on average. Figure \ref{mimic_plot} \textbf{A} shows the proportion of explanations below a threshold, and the vertical line represents the distance value of a change that has occurred once in our train set. With this in mind, we consider counterfactuals below that threshold to be plausible for the purposes of this experiment. We can see that not only are explanations less plausible on average in \textit{ALL}, the proportion of individuals with plausible explanations is significantly lower. Less than a third of individuals receive at least one counterfactual that contains changes we've seen at least once. Consequently, the vast majority of individuals in our test set do not receive a plausible counterfactual if we consider changing all features. Constraining to vital signs, however, leads to only plausible counterfactuals. 

Though we have seen that constraints can be important in improving plausibility, it is important to consider the effect constraints have on the ability of DiCE (or any method for that matter) to generate counterfactuals. Notably, constraints may prevent the generation of valid counterfactuals by not allowing changes to important features. While we were able to generate counterfactuals for all 229 individuals in the test set with \textit{ALL}, we only generated counterfactuals for 120 individuals with \textit{VITAL}. This disparity raises concerns around the tension between plausibility and validity; we can improve plausibility by constraining our search space, but we may constrain counterfactuals in a way that degrades validity.

\subsection{Generating Explanations for Simulated Adult-Income}

\subsubsection{Setup}
Next, we evaluate the use of longitudinal data in the generation process. To maintain comparability between approaches, we compare the same genetic algorithm with different objectives. In the 'Default' algorithm, we optimize sparsity and proximity, and in the 'Longitudinal' algorithm, we optimize proximity and longitudinal distance. 

We use Adult-Income, a common dataset used in fairness and explainability research \cite{adult} \cite{retiring}. The task is to use an individual's demographic and economic information to predict if their income is above 50k. To augment our experiment, we also consider a threshold of 30k. Having two different thresholds should help us understand how plausibility interacts with the rarity of a desired decision. In Adult-Income, 24 percent of individuals have an income above 50k, compared to 44 percent who have an income above 30k. We expect that the lower threshold will lead to more plausible counterfactuals and higher validity for both the 'Default' and 'Longitudinal' methods. 

Though the dataset does not contain any longitudinal data, it is simple enough to reason about what data subjects might look like over time. Therefore, we conduct a simple simulation to generate longitudinal data: we randomly allow some individuals to swap careers with someone else in their education class. We also allow some individuals to increase their level of education before moving to a new career. When swapping careers, all non-demographic variables are swapped (hours-per-week, occupation, and capital loss/gain). Finally, the simulation increases age randomly between one and ten years. This simulation shows some of the ways people can change their economic conditions without allowing any changes on immutable features, such as race or nationality. However, some features we do not include, such as marital status, can change in practice. We focus on allowing changes to features that could potentially be included in a recommendation. 

We train a gradient boosting model and generate ten counterfactuals for individuals in the test set who are predicted to make less than the specified threshold. For our longitudinal distance metric, we normalize continuous features using the MAD, and we normalize categorical features by their observed frequency of change. 

Our primary goal for this experiment is to compare plausibility and validity between 'Longitudinal' and 'Default' approaches to counterfactuals. We change only which metrics are minimized between both experiments. The 'Default' algorithm serves as a naive approach to generating counterfactuals, so only proximity and sparsity are penalized. We do not constrain which features can be changed to elucidate the types of explanations that are produced without knowledge of the data context. 

\subsubsection{Results}

We begin by looking at some summary metrics for the generated counterfactuals, listed in Table \ref{adult_metrics}. We find that 'Default' produces more valid results but more implausible results as well. The vast majority of explanations from 'Default' are valid for both thresholds, but a large majority also contain changes to immutable features, such as marital status and nationality. 'Longitudinal' on the other hand, generated no such counterfactuals. Additionally, the validity gap between 'Default' and 'Longitudinal' is small when the threshold is 30k, but the gap dramatically increases with a threshold of 50k. This signals that plausible and valid counterfactuals are much more achievable when the desired decision is more common. However, validity and rarity of the desired decision are not directly proportional, and their relationship likely varies significantly based on which features are most important in a model. 

\begin{table}[h]
    \centering
    \begin{tabular}{|c||c|c|c|c|}
        \hline
          & 
         \multicolumn{2}{|c|}{30k} & 
         \multicolumn{2}{|c|}{50k} \\ \hline
         \textbf{Metric} & Default & Longitudinal & Default & Longitudinal \\ \hline
         Mean Validity & 0.96 & 0.90 & 0.93 & 0.53 \\ \hline
         \% validity=0 & 0 & 3 & 2 & 22 \\ \hline
         \% validity=1 & 75 & 65 & 72 & 31 \\ \hline
         \% immutable & 73 & 0 & 84 & 0 \\ \hline
    \end{tabular}
    \caption{Metrics for 'Default' and 'Longitudinal' methods across both thresholds. \textit{Validity} refers to whether or not a counterfactual is of the desired class. Since each individual in the test set receives ten counterfactuals, we calculate the mean validity across the those individuals. The last three rows refer to the percent of individuals who have no valid counterfactuals, the percent who have ten counterfactuals, and the percent of counterfactuals that change an immutable feature.}
    \label{adult_metrics}
\end{table}

Next, we look at a few examples of generated explanations to understand how the above summaries play out in practice. Table \ref{adult_example} shows the top two explanations from 'Default' and 'Longitudinal' for two individuals (Subject A and Subject B) who are predicted to make less than 50k. First, we find that Subject A has no valid counterfactual from the 'Longitudinal' method, while Subject B has valid counterfactuals for both methods. Additionally, only one of the 'Default' counterfactuals for Subject B changes an immutable feature, compared to both for Subject A. This is largely due to the fact that female individuals and black individuals make less on average in Adult-Income, and our model hones in on these disparities to make predictions. Consequently, it is more difficult for Subject A to cross the decision boundary without changing some immutable feature. 

\begin{table}[h]
    \centering
    \footnotesize
    \begin{tabular}{c|c c c c c c c c c c | c}
          & Age & HrsWk & Workclass & Edu & Marital & Relationship & Race & Gender & Native-Cntry & Occupation & Prediction \\
         \hline
         Sub. A & 33 & 40 & Fed-gov & Some-col & Never & Not-in-fam & Black & Female & US & Exec & \textcolor{orange}{<50k} \\
         \hline
         Def. & 38 & - & - & - & Married & Husband & White & - & - & - & \textcolor{blue}{>50k}\\
         & 39 & - & - & - & Married & Wife & - & - & Scotland & - & \textcolor{blue}{>50k} \\
         \hline
         Long. & 35 & - & - & - & - & - & - & - & - & Transport & \textcolor{orange}{<50k} \\
         & 37 & 50 & - & - & - & - & - & - & - & Transport & \textcolor{orange}{<50k} \\ \hline\hline

          Sub. B & 37 & 55 & State-gov & HS & Married & Husband & White & Male & US & Protect-serv & \textcolor{orange}{<50k} \\
         \hline
         Def. & - & 54 & Self-emp & - & - & - & - & - & - & Sales & \textcolor{blue}{>50k}\\
         & - & 54 & - & - & - & - & - & - & Taiwan & Exec & \textcolor{blue}{>50k}\\
         \hline
         Long. & 53 & 48 & - & Assoc-voc & - & - & - & - & - & - & \textcolor{blue}{>50k} \\
         & 46 & 56 & - & Bachelors &  - & - & - & - & - & Armed-forces & \textcolor{blue}{>50k} 
         
    \end{tabular}
    \caption{The top explanations for two individuals for both the 'Default' and 'Longitudinal' methods.}
    \label{adult_example}
\end{table}

\section{Discussion} \label{dis}

The results of our work offer some opportunities for improving plausibility in counterfactual explanations, but also some significant concerns with using counterfactuals for recourse. In this section, we discuss some of the implications of Section \ref{exp} along with broader analysis of the use of counterfactual explanations.

\subsection{The Challenge of Recourse}

Our work emphasizes existing critiques of counterfactual explanations and calls into question some foundations of the methodology. Particularly, we emphasize work that has discussed the philosophical and practical difficulties of using counterfactuals for recourse \cite{assumptions}. With this in mind, it is important to ask: are we using explainability for justification or for recourse? The latter is often attainable with existing methodology, since we simply need to show why a decision is justified under our understanding of a model. The former, however, cannot be achieved with purely technical solutions and likely will vary between domains. With recourse, we must think rigorously about what is useful for data subjects, not what the model thinks is most important for a prediction. These are drastically different prerogatives for explainability and should lead to different methodology.

With the goal of recourse, we may use plausibility and achievability as frameworks, so we discuss some of the difficulties of both. Our results show that plausible counterfactuals are difficult and sometimes impossible to attain. While this is partly due to gaps that can be addressed through further research, a significant cause is the underlying data used to train models. Models may hone in on features that are difficult to change or entirely immutable. When that occurs, there may be no plausible change that an individual can make to achieve a desired decision. If a model learns to make predictions, directly or indirectly, based on some class characteristic, an entire class of individuals may be locked out of plausible counterfactuals. Our experiments indicate that diversity of counterfactuals can improve overall plausibility by offering multiple different paths, but diversity alone cannot guarantee plausibility. 

Furthermore, even when we can discover plausible counterfactuals, they may not be achievable. In our experiments with Adult-Income, for example, we found that changing occupation was a common recommendation. While changing careers is technically plausible, it may often be unachievable due to educational, geographic, and personal constraints. That is to say, plausibility is insufficient for recourse. In addition to the mathematical and computational work of designed algorithms for counterfactual explanations, there is significant work to be done regarding the philosophical and practical implementation of counterfactuals for recourse. Elsewise, counterfactual explanations may not become effective tools for recourse.  

Though we have outlined some limitations of counterfactuals, we find that longitudinal data can improve plausibility. Longitudinal data can assist in constraining the counterfactual search space, especially for preventing changes in variables that should not be changed. For features that can be changed, we can constrict counterfactuals to the range of prior observed changes. 

\subsection{Future Work}
In this paper, we only consider a longitudinal distance metric. Future work should explore modeling longitudinal data to further improve plausibility constraints. Future work should also consider implementing intermediate steps across time, and modeling plausiblity in terms of the subject's current features. Additionally, our approach is computationally complex due to row-wise comparison of matrices. Further work can investigate decreasing this complexity, potentially using prototypes or other clustering methods. 
 
\bibliographystyle{ACM-Reference-Format}
\bibliography{main}

\end{document}